%% file: main.tex
\documentclass[10pt,twocolumn,letterpaper]{article}

\usepackage[pagenumbers]{wacv} 

\usepackage{graphicx}
\usepackage{amsmath}
\usepackage{amssymb}
\usepackage{booktabs}
\usepackage{times}
\usepackage{epsfig}
\usepackage{graphicx}
\usepackage{amsmath}
\usepackage{amssymb}
\usepackage{float}
\usepackage{mwe}
\usepackage{multirow}
\usepackage{diagbox}
\usepackage{enumitem}
\usepackage[rightcaption]{sidecap}

\usepackage{pifont}
\usepackage[normalem]{ulem}
\usepackage[accsupp]{axessibility}  
\useunder{\uline}{\ul}{}

\usepackage[pagebackref,breaklinks,colorlinks]{hyperref}

\usepackage[capitalize]{cleveref}
\crefname{section}{Sec.}{Secs.}
\Crefname{section}{Section}{Sections}
\Crefname{table}{Table}{Tables}
\crefname{table}{Tab.}{Tabs.}

\def\wacvPaperID{2072} 
\def\confName{WACV}
\def\confYear{2024}

\def\etc{etc\onedot}
\def\ie{i.e\onedot}
\def\eg{e.g\onedot}
\def\cf{cf\onedot}

\newcommand{\cmark}{\ding{51}}%
\newcommand{\xmark}{\ding{55}}%

\begin{document}

\title{OmniVec: Learning robust representations with cross modal sharing}

\author{Siddharth Srivastava, Gaurav Sharma\\
TensorTour Inc.
\\
{\tt\small \{siddharth, gaurav\}@tensortour.com}
}
\maketitle

\input{abstract}
\input{introduction}
\input{relatedworks}
\input{approach}
\input{experiments}
\input{conclusion}
\appendix
\input{supp}

{\small
\bibliographystyle{ieee_fullname}
\bibliography{main}
}

\end{document}

%% file: abstract.tex
\begin{abstract}
Majority of research in learning based methods has been towards designing and training networks for specific tasks. However, many of the learning based tasks, across modalities, share commonalities and could be potentially tackled in a joint framework. We present an approach in such direction, to learn multiple tasks, in multiple modalities, with a unified architecture. The proposed network is composed of task specific encoders, a common trunk in the middle, followed by task specific prediction heads. We first pre-train it by self-supervised masked training, followed by sequential training for the different tasks.
We train the network on all major modalities, e.g.\ visual, audio, text and 3D, and report results on $22$ diverse and challenging public benchmarks. We demonstrate empirically that, using a joint network to train across modalities leads to meaningful information sharing and this allows us to achieve state-of-the-art results on most of the benchmarks. We also show generalization of the trained network on cross-modal tasks as well as unseen datasets and tasks.
\end{abstract}

%% file: introduction.tex
\section{Introduction}
\label{sec:intro}

Many applied machine learning methods aim to extract useful representations from data. However, a majority of such methods are modality and task specific. Building methods that can work with multiple modalities is a relatively recent research direction~\cite{zhu2022uni,recasens2023zorro, girdhar2022omnivore,li2023uni, jaegle2021perceiver, jaegle2021perceiverio}. Learning tasks together with a unified network can lead to regularization effects, as a large amounts of shared parameters are trained to perform varied tasks, and hence are more likely to extract meaningful representations from data without overfitting to one task or modality. It can also aid in utilizing available labelled data from different domains, hence potentially eliminating the cost and effort of labelling large amounts of data in a specific modality for a specific task. With the ability of sharing knowledge from multiple modalities (\eg image, video, depth map and speech) from different domains (\eg visual, acoustic, textual), the modality agnostic learning frameworks have been shown to provide better robustness~\cite{akbari2021vatt, gong2022uavm} to traditional unimodal networks. We contribute to that line of work, and develop a framework that can learn embeddings in a shared space from different modalities and also deliver high generalization performance. Specifically, we propose to learn embeddings from distinct modalities with modality specific encoders, and process them with a shared transformer backbone. The transformer backbone maps the input embeddings to a shared embeddings space. The network is then trained in an end-to-end manner. 

Prior works towards generalized modality agnostic learning can be categorized into following three approaches. (i) Methods which directly take multiple heterogeneous modalities (image, 3D, audio) as input, without separate encoders for each modality, and directly learn representations from them~\cite{jaegle2021perceiverio, jaegle2021perceiver}. (ii) Methods that take representations from modality specific encoder as input and learn generalized modality specific embeddings using a common objective in the latent space~\cite{baevski2022data2vec}, and, (iii) Methods which aim at sharing knowledge among different modalities by keeping either a common encoder ~\cite{girdhar2022omnivore}  or separate encoders~\cite{akbari2021vatt}. The first two approaches generally target modality agnostic input representation, which lend them capability to keep the network definition same for different modalities. However, such networks, in general, can be trained on one modality at a time, and hence do not facilitate cross modal knowledge sharing. On the other hand, the third approach facilitates jointly training networks on multiple modalities. Our work is closer to the third set of approaches. Specifically, similar to~\cite{baevski2022data2vec}, the proposed method employs different encoder for each modality. 
Similar to~\cite{girdhar2022omnivore} we share knowledge among modalities, and train on multiple modalities sequentially allowing embeddings to generalize across modalities. Unlike~\cite{girdhar2022omnivore}, we do not limit our method to a specific subset of modalities. and train on multiple modalities in a sequential manner. Further, we do not assume any correspondence between the training data \ie paired training sets across modalities, which is different from previous works, \eg~\cite{akbari2021vatt}, where correspondence in data among modalities is assumed. 

Our proposed framework, OmniVec, consists of the following components: (i) a modality specific encoder, (ii) a shared backbone network, and (iii) task specific heads where tasks can be any machine learning task. The framework facilitates end-to-end training. In simple terms, OmniVec works as follows. For a given task and a modality we select a modality compatible encoder and an appropriate task head. We attach the encoder and task heads to, the beginning and end of the shared backbone network respectively. Then to train on another modality, we replace the encoder while keeping the backbone same. If the task is to be changed as well, we replace the task head. To further facilitate learning of better representations and cross-modal information sharing, we train the network numerous tasks. We borrow the motivation from earlier works where it has been shown that training networks on multiple related tasks can provide better generalization~\cite{zhang2021survey}. Similar improvements, in generalization, have been reported for multi-modal multi-task learning as well~\cite{dai2022one, hu2021unit, pramanik2019omninet, srivastava2023hierarchical}. However, we do not train in a traditional multi-task setting, where all tasks are available together and are trained for together. Instead, we train the network in a sequential manner, \ie we train on different tasks, one after another. 

Motivated by empirical observations and previous works indicating that robustness of multi-task mechanisms depends on the complexity of tasks selected for joint training~\cite{ pramanik2019omninet, standley2020tasks}, we propose to group the tasks based on the extent of information exploited by the task across different modalities, \eg, a semantic segmentation task forces the network to embed more local information in the learned representation, as compared to a classification task~\cite{cheng2021per}. In addition to grouping the tasks, we also construct training data by mixing samples from each modality for a particular task. We train the network by replacing modality encoder for each modality, while keeping the task heads and backbone network same. Based on earlier works indicating that self-supervised pretraining helps networks in better exploiting multiple modalities~\cite{dai2022one, girdhar2022omnimae}, we pretrain the network with masked pretraining.

In summary, we make the following contributions.
(i)~We propose a novel method to learn embeddings from many modalities. The method has a common backbone to process the different modalities and perform different tasks. Specifically, we show that the proposed method works with RGB images and videos, depth images, point clouds, audio, speech and text data.
(ii) We propose a novel training mechanism to allow learning using multiple tasks from both spatial (\eg image, 3D point clouds, depth maps) and temporal (\eg video, audio, speech, text) data. Owing to the common backbone of the method, and a synchronous training mechanism, the method shares knowledge between different modalities and tasks, resulting in improved performance and generalization. 
(iii) The proposed method allows for infusing cross domain information in the feature vectors, \ie allowing embeddings from text data to be close to similar data in image domain. 
(iv) We propose an iterative training mechanism by mixing modalities and grouping tasks. Different from earlier works, we also propose to perform self supervised masked pretraining across visual as well as non visual modalities. 
(v) With exhaustive experiments on numerous popular benchmarks across, we show that the proposed framework achieves state-of-the-art results or performs close to the competing methods. (vi) We also study the generalization ability of the proposed framework by demonstrating the robust performance of the learned embeddings on unseen tasks. (vii) We conduct an extensive ablation study to demonstrate the impact of the design choices. 

%% file: relatedworks.tex
\section{Related Works}
\label{sec:relatedworks}

In this section, we discuss similar works and various similar paradigms to our work. We begin with transformers, which are basis of our work, and then move to methods which work with multiple modalities. Among methods that work with multiple modalities, many of them work on utilizing the modalities simultaneously, while others propose networks which take the modalities as input, one at a time.

\par \noindent \textbf{Transformers.} Transformers were proposed originally for Natural Language Processing tasks~\cite{vaswani2017attention}. The main contribution of this work was to demonstrate the effectiveness of multi-head attention in representing long-range correlation between words. Owing to the popularity of transformers in NLP tasks~\cite{lin2022survey}, attempts were made to extend it to vision tasks. Early work in this direction~\cite{yang2020learning, chen2020uniter, luo2020univl} involved utilizing features from convolutional neural networks. However, with vision transformers~\cite{dosovitskiy2020image}, transformers obtained an ability to process raw images and achieved performance competitive to CNNs. After that, transformers have dominated nearly all the vision related tasks~\cite{khan2022transformers}. As transformers have demonstrated robust performance across modalities, recent methods across various modalities use them to solve various tasks~\cite{xu2022transformers, lin2022survey, shamshad2022transformers,selva2023video, fournier2021practical}

\par \noindent \textbf{Multi-modal methods.} Majority of the current multimodal methods use modality specific feature encoders \cite{jiang2021review, kaiser2017one, arandjelovic2018objects, xiao2020audiovisual, recasens2023zorro} and are hence concerned with methods of feature fusion with their proposed architectures. In general the networks for different modalities differ from each other and can not be easily used together without architectural modifications. They also need to decide on when to fuse the features from various modalities, when to fine-tune, how to pre-train \etc~\cite{xu2022multimodal}. Such problems inhibits extending networks such as transformers to be applied as a common backbone across multiple domains such as point clouds, audio and images.
\par \noindent \textbf{Common network for multiple modalities.} Recently, many methods have been proposed which learn from multiple modalities~\cite{ carreira2022hierarchical, girdhar2022omnivore, baevski2022data2vec, jaegle2021perceiver}. Among the most popular, however recent, are methods that do not have separate encoders for each modality. Such methods generally transform the input raw data to a common input representation prior to generally being processed by a transformer network. Among them, the perceiver and similar methods~\cite{jaegle2021perceiver, jaegle2021perceiverio, carreira2022hierarchical} have tried to learn from multiple modalities together without separate encoders. 
Perceiver architecture works by cross-attention among a set of latent queries. Similarly, hierarchical perceiver~\cite{carreira2022hierarchical} builds upon it proposes to group the input array while preserving the locality structure. On the other hand, methods such as data2vec~\cite{baevski2022data2vec} use modality specific encoders. Other methods such as Omnivore~\cite{girdhar2022omnivore} have a common encoder. However, Omnivore is limited to only visual modalities (image, depth map, video). Then, methods such as VATT~\cite{akbari2021vatt} have a common backbone for each text, image and audio. However, it processes each modality independently using a transformer. Such methods which learn from multiple modalities have been shown to provide better robustness ~\cite{akbari2021vatt, gong2022uavm}. Our methods largely overlaps with the motivation of such methods, however, it differs from such methods in that earlier methods operate on training for one task or one modality at a time, while we learn by training on multiple modalities and multiple tasks while using a single common backbone architecture.
\par \noindent \textbf{Multi-task learning.} We have discussed many methods that attempt at learning from multiple inputs. As discussed in the previous section, recent years have seen many methods that work with multiple modalities. PerceiverIO~\cite{jaegle2021perceiverio} extends Perceiver~\cite{jaegle2021perceiver} and enables learning multiple tasks using the same network architecture. While PerceiverIO can also learn multiple tasks at a time using a single architecture, generally multiple networks are used~\cite{zhang2018overview}. Many techniques~\cite{baevski2022data2vec, girdhar2022omnivore, hu2021unit, pramanik2019omninet, dai2022one} learn from multiple modalities and from their raw representation and apply to multiple tasks. 
\par \noindent \textbf{Multi-modal masked pretraining.}
Methods such as ~\cite{liu2021opt, yan2022multi, wei2022masked} use masked pre-training.
Masked pretraining has shown to improve the performance of deep networks networks for various modalities and tasks~\cite{akbari2021vatt, girdhar2022omnimae, baevski2022data2vec, baade2022mae, yu2022point, gupta2022maskvit} and motivated by such works we also use masked pre-training as a self supervised step leveraging large amounts of data available. However, different from earlier works, we perform masked pre-training on multiple modalities and multiple datasets on the same common backbone. 

%% file: approach.tex
\section{Approach}
\label{sec:approach}

\begin{SCfigure*}[0.75][ht]
    \centering
    \includegraphics[width=0.75\textwidth]{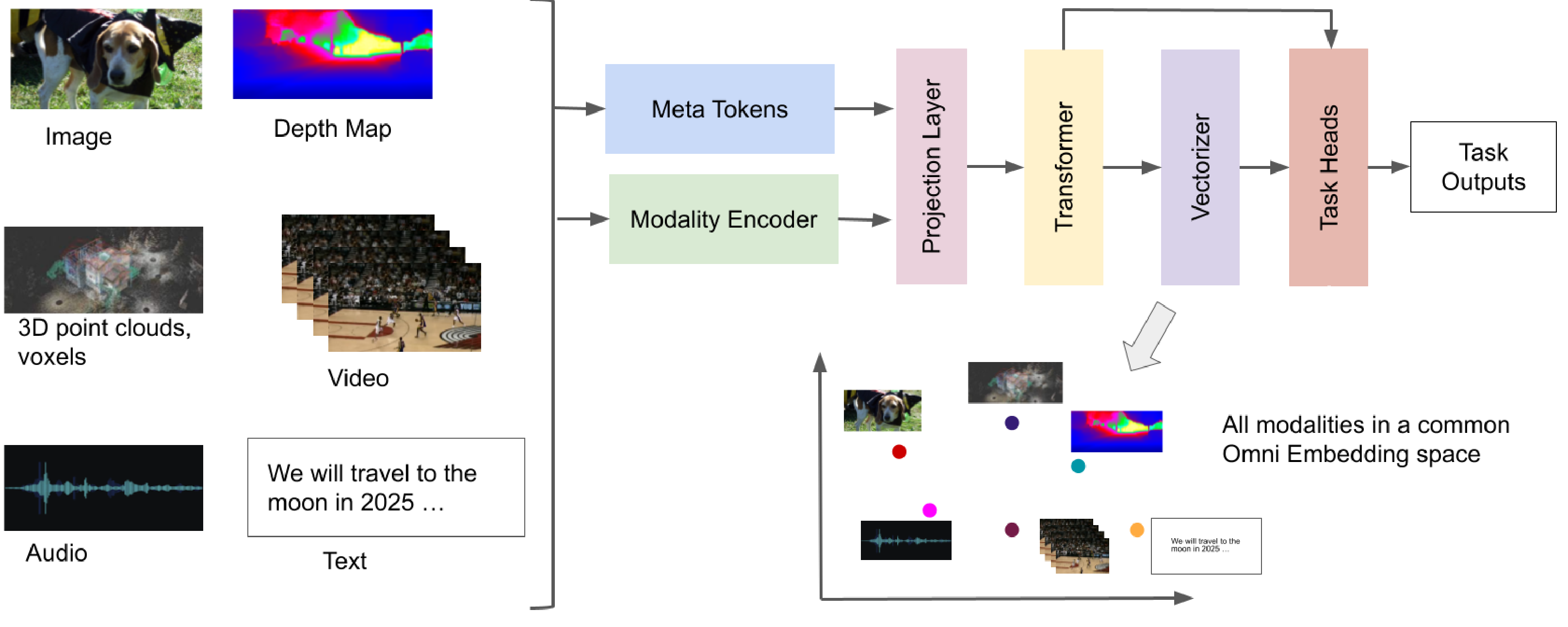}
    \caption{\textbf{OmniVec:}
    The proposed method takes data from one of the modalities and pass it through the modality encoder and combine it with the meta token and then pass through the projection layer to embedd the feature onto a common embedding space. Then it is passes through the common backbone of Transformer layers which is then vectorized by the vectorizer. Finally, the task heads are used for task specific outputs.}
    \label{fig:architecture}
\end{SCfigure*}

We now describe our framework for learning multiple tasks in multiple modalities with a common backbone network, allowing for cross modality knowledge sharing. The overview of the proposed framework is shown in Figure \ref{fig:architecture}. The network comprises six building blocks, \ie modality encoders, meta token block, projection block, transformer, vectorizer and task heads. We now explain each block in detail.

\subsection{OmniVec Framework}
\noindent\textbf{Modality Encoder.}  The modality encoder takes as input, one modality at a time and extracts feature embedding for each of the modalities. In the proposed framework, the modality encoder can be a transformer, convolutional neural network or can directly use raw signals~\cite{akbari2021vatt}.  As we do not assume any specific structure for the modality encoder, the proposed framework allows incorporating any appropriate deep network as a modality encoder. 

For current work, we use domain specific transformer based encoders for each of the modalities as shown in Table \ref{tab:modality_encoder} followed by a common backbone network. It is worth noting that each of the networks in visual and auditory domain is based on Vision Transformer architecture \ie image and depth directly use ViT, video (ViViT) differs from ViT in input tokenization that extends 2D patches to 3D (spatio-temporal mapping), audio (AST) transformers differ from ViT only in input representation \ie uses log-mel spectrograms instead of images, Simple3D-former for point cloud uses a 2D ViT transformer as the base network with modified positional embeddings and tokenization approach. We use a standard BERT transformer for textual data. We train each of these models from scratch. 

\begin{table}[]
\resizebox{\columnwidth}{!}{ 
\begin{tabular}{@{}lll@{}}
\toprule
\textbf{Modality} & \textbf{Domain} & \textbf{Network}                                                       \\ \midrule
Image             & Visual          & Vision Transformer (ViT)~\cite{dosovitskiy2020image}                                                \\
Depth maps        & Visual          & Vision Transformer (ViT)~\cite{dosovitskiy2020image}                                                \\
Video             & Visual          & Video Vision Transformer (ViViT)~\cite{arnab2021vivit}                                                              \\
3D point clouds   & Visual          &  Simple3D-former~\cite{wang2022can}                                          \\
Audio            & Auditory           & Audio Spectrogram Transformer (AST)~\cite{gong2021ast} \\
Text              & Language            & BERT~\cite{devlin2018bert}                                                                   \\ \bottomrule
\end{tabular}
}
 \caption{\textbf{Modality Encoders.} We select transformer based modality encoders for evaluating OmniVec framework}
 \label{tab:modality_encoder}
\end{table}

\noindent\textbf{Meta Tokens.} We extract meta tokens from the input modalities. This meta representation is a vector that encodes the type of modality ($I$), size of temporal dimension ($T$), height ($H$), width ($W$) in spatial dimension, number of channels ($C$) and length or number of tokens ($L$). In general, the meta tokens can also hold additional information to make the framework adapt to additional modalities.  The value in each of these representation variables is conditioned on the type of modality \eg non spatial data may have $H$, $W$ only with the other non-spatial parameter set as a special token, denoting lack of such information. 
    
\noindent\textbf{Projection Layer.} The projection layer inputs the intermediate representations from the modality encoder network and is conditioned on the meta tokens. It then converts the input representation to patches that are provided as input to the subsequent transformer network. We obtain $n$-dimensional vector for each patch by applying linear projection. Similar to ViT~\cite{dosovitskiy2020image}, this projection is applied with a learnable weight $W_{ip} \in \mathbb{R}^{t\cdot h \cdot w \cdot c \cdot l \times n}$ for each modality $i$.  The meta tokens make the projection layer adaptable to varying number and dimensions of input patches and generate latent representations compatible with the subsequent transformer network. For instance, we represent RGB images as $I \in \mathbb{R}^{1 \times h \times w \times 3 \times 1}$ with $t$=$1$ frames and $c$=3 channels. Similarly, we represent video as $V \in \mathbb{R}^{t \times h \times w \times 3 \times 1}$ with $t$ frames ($t > 1$) and $c=3$ channels, depth as $D \in \mathbb{R}^{1 \times h \times w \times 4 \times 1}$ with $c=4$ channels, point cloud as $P \in \mathbb{R}^{1 \times 1 \times 1 \times 3 \times l}$ with $l$ points, audio as $A \in \mathbb{R}^{t \times h \times w \times c \times 1}$ with spectrogram input, and text as $L \in \mathbb{R}^{1 \times 1 \times 1 \times 1 \times l}$ with $l$ tokens. Each patch $\mathbf{x}$ is processed independently and projected to an embedding $\mathbf{e}$ followed by a LayerNorm~\cite{ba2016layer}

\noindent\textbf{Transformer.} The transformer network is the common part of the framework and is in effect a `bottleneck` block. While different modalities may arrive here through different encoders, they all have to pass through this transformer network. The transformer network inputs the patches generated by the projection layer and outputs features. While the OmniVec framework can use any standard transformer architecture, we use ~\cite{devlin2018bert} as our backbone architecture. In our transformer network, the multi head attention involves standard self-attention~\cite{vaswani2017attention},  and GeLU~\cite{hendrycks2016gaussian} activation prior to the MLP layer.

\noindent\textbf{Vectorizer} The vectorizer layer takes patches from the transformer network as input, and outputs embeddings for the original data point. It outputs a single embedding $\mathbf{e} = f(\mathbf{X})$ for an input $\mathbf{X}$. We name the output embeddings of the vectorizer as Omni Embeddings, as these embeddings constitute knowledge from multiple tasks and modalities due to forward pass from the transformer block where cross modality and cross task information is infused. 

For our implementation, we concatenate the output patches and pass them through a linear layer to obtain a $d$-dimensional embedding. At the time of training, we use the outcome of vectorizer as input to subsequent task heads. However, using the outcome of vectorizer as input to task heads is optional as the task head may also directly take input patches from the previous transformer bottleneck. Once the model has been trained, the output from vectorizer can be used for fine-tuning and evaluation on downstream tasks. 

\noindent\textbf{Task Heads} The final parts of network, the task heads are $\sum{T_{ih}}$ independent networks which learn task $h$ for every $i^{th}$ modality. The task heads can generally be any computer vision, natural language processing or other modality specific task. We experiment with classification (image, video, audio, text), segmentation (image, point clouds) \etc We describe them in Section \ref{sec:experiments}.

\begin{table*}[]
\resizebox{\textwidth}{!}{ 
\begin{tabular}{llm{4em}m{4em}m{3em}m{3em}m{3em}ccccc}
\hline
\textbf{Method/Dataset} & \textbf{Supp. Modalities}                                                                           & \textbf{Cross-Modal sharing} &  \textbf{Masked pretraining}                                                                           &\textbf{Supp. Tasks} & \textbf{AudioSet (A+V.)} & \textbf{AudioSet (A)} & \multicolumn{1}{l}{\textbf{SSv2}} & \multicolumn{1}{l}{\textbf{GLUE}} & \multicolumn{1}{l}{\textbf{ImageNet1K}} & \multicolumn{1}{l}{\textbf{Sun RGBD}} & \multicolumn{1}{l}{\textbf{ModelNet40}} \\ \hline
Omni-MAE~\cite{girdhar2022omnimae}              & Image, Video                                                                                  & No &  Yes &Class.                                              & -                                                   & -                                                 & 73.4                                                & -                                 & 85.5                                    & -                                     & -                                       \\
Perceiver~\cite{jaegle2021perceiver}                                   & Modality Agnostic & No & No                                                                             & Class.                                               & 43.4                                                & 38.4                                              & -                                                   & -                                 & 78.6                                    & -                                     & -                                       \\
Heirarchical Perceiver~\cite{carreira2022hierarchical}                      & Modality Agnostic  & No & No                                                                           & Class.                                               & 43.8                                                & 41.3                                              & -                                                   & -                                 & 81.0                                    & -                                     & 80.6                                    \\
data2vec~\cite{baevski2022data2vec}                                    & Modality Agnostic   & No & Yes                                                                          & Class.                                               & -                                                   & 34.5                                              & -                                                   & 82.9                              & 86.6                                    & -                                     & -                                       \\
Omnivore~\cite{girdhar2022omnivore}                                    & Image, Video, Depth map                                                                       & Yes & No & Class.                                             & -                                                   & -                                                 & 71.4                                                & -                                 & 84.0                                    & 65.4                                  & -                                       \\
VATT~\cite{akbari2021vatt}                                        & Image, Video, Audio, Text                                                                     & Yes  & Yes & Class.                                            & -                                                   & 39.4                                              & -                                                   & -                                 & -                                       & -                                     & -                                       \\ 
Perceiver IO~\cite{jaegle2021perceiverio}                                & Modality Agnostic                                                                             & No & No & Multiple                                              & -                                                   & -                                                 & -                                                   & -                                 & 79.0                                    & -                                     & 77.4                                    \\
\hline
OmniVec (pretrained)                        & \begin{tabular}[c]{@{}c@{}}Image, Video, Audio, Text, \\ Depth map, Point Clouds
\end{tabular} & Yes    & Yes & Multiple                                          & \textbf{48.6}                                       & \textbf{44.7}                                     & \textbf{80.1}                                       & \textbf{84.3}                     & \textbf{88.6}                           & \textbf{71.4}                         & \textbf{83.6}                           \\ \hline
\end{tabular}
}
\caption{ \textbf{Comparison of OmniVec framework with similar methods that work on multiple modalities}. We compare OmniVec with masked pretraining with the best reported results from respective publications of the compared methods. Supp. Tasks and Supp. Modalities indicate Supported Tasks and Supported Modalities by respective networks. In Supported (Supp.) Tasks, Class. indicates classification.}
\label{tab:res_masked_pretraining}
\end{table*}

\subsection{Training OmniVec Framework}
We train the OmniVec Framework in two stages. First we perform masked pretraining. Then we fine tune the network on multiple modalities. Both these stages are described below.

\noindent\textbf{Masked Pretraining.} We pretrain the network with masked autoencoders~\cite{girdhar2022omnimae, akbari2021vatt}. Specifically, for an input with $N$ patches, we mask $K$ patches, and feed non-masked patches and their positions to the encoder. For each modality, we use the encoder from Table~\ref{tab:modality_encoder} followed by our bottleneck transformer that outputs per patch embeddings \ie we keep a shared bottleneck transformer encoder for each of the modalities. Similar to ~\cite{he2022masked, akbari2021vatt}, the per patch embeddings are concatenated with $K$ replicas of learnable mask tokens resulting in $N$ embeddings. We add corresponding positional embeddings to each of the $N$ embeddings, and pass to the decoder. We use the same masking strategy for modalities from visual and auditory domains. For textual data, we follow~\cite{song2020mpnet} and randomly permute the sentences~\cite{yang2021generalized} and use a small fraction $f$ of tokens as predicted tokens, followed by utilizing $8$:$1$:$1$ strategy of BERT~\cite{devlin2018bert} for constructing mask tokens. The training objective is to minimize the reconstruction error between the input and decoder outputs. For image, video, point clouds and audio spectrogram input, we minimize $l_2$ distance between the $K$ predicted and target patches. For visual inputs, the input samples are normalized to zero mean and unit variance. For textual data, we use the permuted language modelling of XLNet~\cite{yang2021generalized} as the objective. 

\noindent\textbf{Training on multiple modalities and tasks.} For training the network on multiple modalities and tasks, we introduce \textit{modality mixing} and \textit{task grouping}. We train our model using a collection of $h$ tasks $\mathbf{T_{i,h}}$ for $i^{th}$ modality. We group tasks into \textit{simple} and \textit{dense} tasks and refer to it as \textit{task grouping}. We categorize the tasks into two categories namely, simple and dense based on the complexity of the dataset and outputs \ie classification task predicts a single label for a given input, irrespective of the size of the input, therefore we refer it as a simple task. However, a segmentation or depth prediction task, requires each pixel to be predicted, and hence we refer it as a dense task. We detail each of the tasks, the datasets used to train them and their task grouping in Section \ref{sec:experiments}. 

As we do not assume any correspondence between data from various modalities, we propose mixing samples from all datasets for a particular task to share knowledge between various modalities. An alternative approach would be to construct mini-batches from each dataset separately. However, we found it performs poorly compared to mixing samples from modalities. We refer this strategy of constructing mini-batches as \textit{modality mixing}. Specifically, for a particular task $h$ belonging to a type of task $t$ (simple, dense), for each modality $i$, we extract sample $s_{t,i,h}$ from the datasets.  

After task grouping and modality mixing, we train the network in an end-to-end manner iteratively for simple and dense tasks. Specifically, we train the network for $E$ epochs, we train the network for $v_1$ epochs with mini-batches from simple tasks and $v_2$ for dense tasks. We continue training the network in iterative manner \ie switching between simple and dense tasks for $E$ epochs.

%% file: experiments.tex
\begin{table*}
\centering
\parbox{0.55\columnwidth}{
\centering
\resizebox{0.55\columnwidth}{!}{ 

\begin{tabular}{cm{3.6em}m{3em}}
\hline
\textbf{Method/Dataset}         & \textbf{iNaturalist 2018} & \textbf{Places 365}                  \\
\hline
Omni-MAE~\cite{girdhar2022omnimae}                       & 78.1            & 59.4                           \\
Omnivore~\cite{girdhar2022omnivore}                        & 84.1            & 59.9                           \\
EfficientNet B8\cite{tan2019efficientnet}                 & 81.3            & 58.6                           \\
MAE\cite{he2022masked}                             & 86.8            &                                \\
MetaFormer~\cite{yu2022metaformer}                      & 87.5            & 60.7                           \\
InternImage\cite{wang2023internimage} & {\ul 92.6}   & {\ul 61.2} \\
\hline
OmniVec                         & \textbf{93.8}      & \textbf{63.5}  \\
\hline
\end{tabular}}
\caption{\textbf{iNaturalist-2018 and Places-365} top-$1$ accuracy. 
}
\label{tab:sota_image}
}
\hspace{0.5em}
\parbox{0.45\columnwidth}{
\centering
\resizebox{0.45\columnwidth}{!}{ 

\begin{tabular}{cc}
\hline
\textbf{Method/Dataset} & \textbf{Kinetics-400} \\
\hline
Omnivore~\cite{girdhar2022omnivore}                & 84.1                  \\
VATT~\cite{akbari2021vatt}                    & 82.1                  \\
Uniformerv2~\cite{li2022uniformerv2} & 90.0 \\
InternVideo\cite{wang2022internvideo}             & \textbf{91.1}         \\
TubeViT\cite{piergiovanni2023rethinking}                 & {\ul 90.9}                  \\
\hline
OmniVec                 & \textbf{91.1} \\\hline                
\end{tabular}}
\caption{\textbf{Kinetics-400} top-$1$ accuracy.
}
\label{tab:sota_kinects400}
}
\hspace{0.5em}
\parbox{0.45\columnwidth}{
\centering
\resizebox{0.45\columnwidth}{!}{ 

\begin{tabular}{cm{4em}}
\hline
\textbf{Method/Dataset} & \textbf{Moments in Time} \\
\hline
VATT~\cite{akbari2021vatt}                    & 41.1                     \\
Uniformer v2\cite{li2022uniformerv2}           & 47.8                     \\
CoCa\cite{yu2022coca}                    & 47.4                     \\
CoCa-finetuned\cite{yu2022coca}          & {\ul 49.0}               \\
\hline
OmniVec                 & \textbf{49.8} \\
\hline
\end{tabular}
}
\caption{\textbf{Moments in time} top-$1$ accuracy.
}
\label{tab:sota_MIT}
}
\hspace{0.5em}
\parbox{0.4\columnwidth}{
\resizebox{0.4\columnwidth}{!}{
    \begin{tabular}{cc}
        \hline
        \textbf{Method/Dataset} & \textbf{ESC50} \\
        \hline
        AST~\cite{gong2021ast}                     & 85.7           \\
        EAT-M\cite{gazneli2022end}                   & 96.3           \\
        HTS-AT\cite{chen2022hts}                  & 97.0           \\
        BEATs\cite{oreshkin2019n}                   & {\ul 98.1}     \\
        \hline
        OmniVec                 & \textbf{98.4} \\
        \hline
    \end{tabular}
}
\caption{\textbf{ESC50} top-$1$ accuracy. 
}
\label{tab:sota_audio}
}
\hspace{1em}
\parbox{0.45\columnwidth}{
\resizebox{0.45\columnwidth}{!}{

}
}
\end{table*}

\section{Experiments}
\label{sec:experiments}

\noindent\textbf{Masked pretraining.} We do masked pretraining using the modality mixing as described in Section \ref{sec:approach}. We use AudioSet (audio) \cite{gemmeke2017audio}, Something-Something v2 (SSv2)(video) \cite{goyal2017something}, English Wikipedia (text), ImageNet1K (image) \cite{deng2009imagenet}, SUN RGB-D (depth maps) \cite{song2015sun}, ModelNet40 (3D point cloud) \cite{wu20153d} for pretraining the network.  As we perform autoencoder based pre-training, we do not group the tasks, and instead uniformly sample data from each of the datasets and modalities. Further, we randomly select patches for masking. For image, video and audio, we randomly mask $90\%$ of the patches. For point cloud, we mask $80\%$ of the patches, and for text we mask $95\%$ of the patches. Further, we keep $f$ = $5\%$ of the tokens as predicted tokens (unlike $15\%$ in \cite{yang2021generalized}). We perform pretraining for $2000$ epochs.  

\noindent\textbf{Modality Encoder.} For modality specific encoders, we use the networks from Table \ref{tab:modality_encoder}. We use the same network configurations for these networks as in corresponding publications. We pretrain the model using masked pretraining as described in Section \ref{sec:approach}, followed by training on specific modalities as per task groups and modality mixing. For different tasks on a modality, we keep the modality encoder same, while changing the task heads with appropriate loss functions. We train modality encoders for $E=900$ epochs with $2$ consecutive epochs each for simple and dense task groups.  

\noindent\textbf{Datasets for training on multiple modalities and tasks.} After masked pre-training, we fine tune the network on multiple tasks across modalities. 
The datasets and their corresponding task groups and modality are given in Table~\ref{tab:multi_task_training}.

\noindent\textbf{Task Heads.} For classification tasks, we use standard classification head from ViT~\cite{dosovitskiy2020image} while use ~\cite{arnab2021vivit} for video classification and ~\cite{gong2021ast} for audio classification.  For image and point cloud segmentation tasks, we use the segmentation head from~\cite{ranftl2021vision}. For text summarization, we use a $3$-layered transformer.

We provide more implementation details in the supplementary material.

\begin{table}[]
\resizebox{\columnwidth}{!}{ 
\begin{tabular}{cccc}
\hline
\textbf{Task}              & \textbf{Dataset}              & \textbf{Modality} & \textbf{Task Group} \\
\hline
Image Recognition          & iNaturalist-2018 \cite{van2018inaturalist}             & Image             & Simple              \\
Scene Recognition          & Places-365 \cite{zhou2017places}                 & Image             & Dense               \\
Video Action Recognition   & Kinetics-400 \cite{kay2017kinetics}                 & Video             & Simple              \\
Video Action Recognition   & Moments in Time \cite{monfort2019moments}              & Video             & Dense               \\
Audio Event Classification & ESC50 \cite{piczak2015esc}                         & Audio             & Simple              \\
Point Cloud Segmentation   & S3DIS \cite{armeni20163d}                        & Point Cloud       & Dense               \\
Text Summarization         & DialogueSUM \cite{chen2021dialogsum}                & Text              & Dense               \\
Point Cloud Classification & ModelNet40-C \cite{wu20153d}                 & Point Cloud       & Simple      \\       
\hline
\end{tabular}
}
\caption{\textbf{List of tasks and corresponding datasets for task group based training after masked pretraining}. We assign each task to a task group (simple, dense) based on complexity of the dataset and output.}
\label{tab:multi_task_training}
\end{table}

\subsection{Results}

\noindent\textbf{Comparison of pretrained OmniVec with similar methods.}
Table~\ref{tab:res_masked_pretraining} compares OmniVec model with masked pretraining to various similar methods. The table also indicates the modalities supported by various methods (Col.-Supp. Modalities), and that if the method supports sharing knowledge between modalities (Col.-Cross-Modal sharing). Further, it also details the learning objectives by these methods. The table reports results on six benchmark datasets on seven tasks as AudioSet supports two tasks (audio only, and audio with video). These datasets are used to perform masked pretraining on the OmniVec model as described in Section~\ref{sec:approach}. It can be observed that the proposed OmniVec model outperforms all the compared methods on all the datasets. It is important to note that, we do not fine tune on any of these datasets specifically while other methods, in general, fine tune the results, mostly using a linear layer with softmax classification. This demonstrates the robustness of the proposed model and its ability to learn generalized embeddings without task specific fine-tuning.

\noindent\textbf{Comparison to state-of-the-art.} For comparison with state of the art methods, we performed masked pretraining of OmniVec followed by training on multiple modalities and task groups as described in Section \ref{sec:approach}. We discuss the comparison on each modality below.

\noindent\textbf{(i) Image}
Table~\ref{tab:sota_image} shows state of the art on image datasets. We compare with multi-modal methods (Omni-MAE, Omnivore) and specialized methods (MetaFormer, InternImage). We surpass the state of the art on iNaturalist with a top-1 accuracy of 93.8\%, compared to InternImage's 92.6\%. On Places-365, we beat all competitors, achieving 61.6\% accuracy versus InternImage's 61.2\%. Moreover, we best Omnivore by  $\sim7$\% on iNaturalist and  $\sim3$\% on Places-365. Our results either match or surpass modality-specific methods in image classification, and outperforming unified learning methods.

\noindent\textbf{(ii) Video} Table~\ref{tab:sota_kinects400} and Table~\ref{tab:sota_MIT} show comparison against state of the art methods on Kinetics-400 and Moments in Time datasets.We observe that we outperform all the competing methods on Moments in Time dataset while perform same as the state of the art method InterVideo  \ie $91.9$ top-1 accuracy.    

\noindent\textbf{(iii) Audio} 
Table~\ref{tab:sota_audio} highlights our comparison with top-performing methods on the ESC50 dataset. OmniVec outperforms competing methods, achieving an accuracy of 98.4\%, significantly higher than the Audio Spectrogram Transformer (AST) at 85.7\%. While most compared methods utilize supervised pretraining on AudioSet, we adopt masked pretraining without accessing labels. This suggests OmniVec's proficiency in learning from related tasks across different modalities, emphasizing its effectiveness in cross-modal knowledge transfer.

\begin{table*}[]
\centering
\resizebox{\textwidth}{!}{ 
\begin{tabular}{lccccccccc}
\hline
\multicolumn{1}{c}{\textbf{Method}} &  
\multicolumn{1}{c}{\textbf{Task Grouping}} & \multicolumn{1}{c}{\textbf{Modality Mixing}} & \multicolumn{1}{l}{\textbf{AudioSet (A+V.)}} & \multicolumn{1}{l}{\textbf{AudioSet (A)}} & \multicolumn{1}{l}{\textbf{SSv2}} & \multicolumn{1}{l}{\textbf{GLUE}} & \multicolumn{1}{l}{\textbf{ImageNet1K}} & \multicolumn{1}{l}{\textbf{Sun RGBD}} & \multicolumn{1}{l}{\textbf{ModelNet40}} \\ \hline

OmniVec-1 (baseline) & \xmark & \xmark & 37.5
& 36.3  
& 62.6
& 57.5
& 70.2
& 59.8 
& 68.5 \\ 
OmniVec-2 & \cmark & \xmark & 42.6 
& 40.1  
& 73.5
& 69.5
& 79.8
& 66.4 
& 75.2 \\ 
OmniVec-3 & \xmark & \cmark & 39.2
& 39.4  
& 70.2
& 68.8
& 77.3
& 65.5 
& 72.2 \\ 
OmniVec-4 & \cmark & \cmark & \textbf{48.6} 
& \textbf{44.7}                                    & \textbf{80.1}                                    & \textbf{84.3}
& \textbf{88.6}
& \textbf{71.4} 
& \textbf{83.6}\\ 

\hline
\end{tabular}
}
\caption{\textbf{Impact of various training strategies on OmniVec.} We report results with and without each of task grouping and modality mixing. The results are reported with masked pretraining only. We observe that individually, both task grouping and modality mixing improve the results over the baseline method. However, there combination outperforms individual performance using these mechanisms.}
\label{tab:ablation}
\end{table*}

\begin{table*}
\centering
\parbox{0.55\columnwidth}{
\centering
\resizebox{0.5\columnwidth}{!}{
\begin{tabular}{cm{4em}}
\hline
\textbf{Method/Dataset} & \textbf{Model Net40C} \\
\hline
PointNet++\cite{qi2017pointnet++}              & 0.236                 \\
DGCN+PCM-R\cite{zhang2022pointcutmix}     & 0.173                 \\
PCT + RSMIx\cite{lee2021regularization}             & 0.173                 \\
PCT + PCM-R\cite{sun2022benchmarking}     & {\ul 0.163}           \\
\hline
OmniVec                 & \textbf{0.156}   \\  
\hline
\end{tabular}
}
\caption{\textbf{ModelNet40-C} Error Rate.
}
\label{tab:sota_mnc}
}
\hspace{1em}
\parbox{0.65\columnwidth}{
\centering
\resizebox{0.5\columnwidth}{!}{
\begin{tabular}{cc}
\hline
\textbf{Method/Dataset} & \textbf{S3DIS} \\
\hline
PointTransformer+CBL\cite{tang2022contrastive}    & 71.6           \\
StratifiedTransformer\cite{lai2022stratified}   & 72.0           \\
PTv2\cite{wu2022point}                    & { 72.6}     \\
Swin3D\cite{yang2023swin3d}                   & {\ul 74.5}     \\
\hline
OmniVec                 & \textbf{75.9}  \\
\hline
\end{tabular}
}
\caption{\textbf{Stanford Indoor Dataset} mIoU.
}
\label{tab:sota_s3dis}
}
\hspace{1em}
\parbox{0.75\columnwidth}{
\centering
\resizebox{0.6\columnwidth}{!}{
\begin{tabular}{ccccc}
\hline
\textbf{Method} & \textbf{R-1}   & \textbf{R-2}   & \textbf{R-L}   & \textbf{B-S}  \\
\hline
CODS\cite{wu2021controllable}            & 44.27          & 17.90          & 36.98          & 70.49         \\
SICK\cite{kim2022mind}            & {\ul 46.2}     & {\ul 20.39}    & \textbf{40.83} & {\ul 71.32}   \\
\hline
OmniVec         & \textbf{46.91} & \textbf{21.22} & {\ul 40.19}    & \textbf{71.91} \\
\hline
\end{tabular}
}
\caption{\textbf{DialogueSUM} text summarization ROGUE scores. 
}
\label{tab:sota_dialoguesum}
}

\end{table*}

\noindent\textbf{(iv) Point Cloud.} Table~\ref{tab:sota_mnc} and Table~\ref{tab:sota_s3dis} compare against state of the art methods on ModelNet40-C and S3DIS datasets respectively. On ModelNet40-C, we evaluate a classification task, while on S3DIS we evaluate semantic segmentation. On both the datasets, we outperform the competing method. This demonstrates that the proposed method is able to robust performance with the shared backbone network across tasks. 

\noindent\textbf{(v) Text} 
Table~\ref{tab:sota_dialoguesum} shows state of the art on DialogueSUM dataset for text summarization. OmniVec surpasses other methods in three out of four metrics and comes in second on the R-L metric. Despite utilizing significantly fewer datasets for text (only two) in comparison to visual tasks (ten datasets), OmniVec demonstrates strong performance. This suggests OmniVec's capacity to bridge the modality gap~\cite{liang2022mind} across distinct domains in the latent space, even when the data distribution is skewed.

\subsection{Ablations}

\noindent\textbf{Impact of task grouping and modality mixing.} Table~\ref{tab:ablation} shows the effect of task grouping and modality mixing. We evaluate four network variations: (i) OmniVec-1 without either of task grouping and modality mixing, (ii) OmniVec-2 with just task grouping, (iii) OmniVec-3 with only modality mixing, and (iv) OmniVec-4 combining both. OmniVec-1 uses masked pretraining on single datasets. OmniVec-2 groups tasks by modality, OmniVec-3 mixes modalities randomly, and OmniVec-4 follows the settings from Section \ref{sec:approach}. Comparatively, OmniVec-1 lags behind the others. Both OmniVec-2 and OmniVec-3 outperform OmniVec-1 by around 30\% to 45\%, showing their efficacy. However, OmniVec-4, which combines both approaches, performs better, emphasizing the benefits of integrating tasks and modalities.

\begin{table*}
\centering
\resizebox{\textwidth}{!}{ 
\begin{tabular}{ccccccc}
\hline
\textbf{Dataset} & \textbf{Modality} & \textbf{Task} & \textbf{Metric}  & \textbf{OmniVec (Pre.)} & \textbf{OmniVec (FT.)}  & \textbf{SOTA}        \\
\hline
UCF-101 &  Video & Action Recognition  &   3-Fold Accuracy  & {\ul 98.7} & \textbf{99.6}  &   \textbf{99.6} (VideoMAE V2-g\cite{wang2023videomae}) \\
HMDB51 & Video  & Action Recognition  &   3-Fold Accuracy  & {\ul 89.21} & \textbf{91.6}  &   88.1 (VideoMAE V2-g\cite{wang2023videomae}) \\
Oxford-IIIT Pets &  Image & Fine grained classification  &   Top-1 Accuracy  & \uline{97.4} &  \textbf{99.2} &   97.1 (EffNet-L2\cite{foret2020sharpness}) \\
ScanObjectNN   & 3D Point Cloud & Classification  &   Accuracy  & 92.1 & \textbf{96.1}  &   \uline{93.4} (PointGPT\cite{chen2023pointgpt}) \\
NYU V2 & RGBD  & Semantic Segmentation  &   Mean IoU  & \uline{58.6} &  \textbf{60.8} &   56.9 (CMN\cite{liu2022cmx}) \\
SamSum & Text  & Meeting Summarization  &   ROGUE(R-L)  & \uline{51.2} &   \textbf{54.6} &   50.88 (MoCa\cite{zhang2022momentum}) \\
KITTI & RGB & Depth Prediction  &   iRMSE  & - & \textbf{10.2}  &   \uline{10.4} (VA-DepthNet\cite{liu2023va}) \\
YouCook2 & Video+Text  & Zero Shot Text-to-Video Retrieval  &   Recall@10  & \uline{64.2} &  \textbf{70.8} &   63.1 (VideoCLIP\cite{xu2021videoclip}) \\
MSR-VTT & Video+Text  & Zero Shot Text-to-Video retrieval  &   Recall@10  & 78.6 & \uline{89.4}  &  80.0(Pre.)/\textbf{90.8}(FT)(SM\cite{zeng2022socratic}) \\
\hline
\end{tabular}
}
\caption{\textbf{Generalization performance of OmniVec} on \textit{unseen datasets} (Oxford-IIIT Pets, UCF-101, HMDB51, ScanObjectNN, NYUv2 Seg, SamSum), \textit{unseen tasks} (KITTI Depth Prediction) and \textit{cross-domain} generalization (YouCook2, MSR-VTT). Pre. indicates network with pretraining only, FT indicates network finetuned on training set of respective datasets. See supplementary for more detailed results.}
\label{tab:generalization}
\end{table*}

\noindent\textbf{Influence of size of the modality encoder.} 
We evaluated the impact of enlarging the base modality encoder to the scale of our suggested network, using modality-specific data. This change slightly improved performance. For example, on ImageNet1K, the top-1 accuracy went from 88.5\% with the base ViT~\cite{dosovitskiy2020image} to 89.1\% with the augmented ViT having a similar parameter count, while OmniVec achieved 92.4\%. These findings suggest that even with enhancements, the augmented base modality encoder still lags significantly behind OmniVec, highlighting OmniVec's advantage of leveraging information from multiple modalities.

\noindent\textbf{Fine-tuning with the same datasets after masked pretraining and comparison to state-of-the-art.} In Table~\ref{tab:sota}, we show the results of fine-tuning the OmniVec-4 model on each of the datasets that was used for masked pretraining. As during masked pretraining, we use the standard train sets for each of these datasets for fine-tuning. 

It can be observed from the results that OmniVec achieves better performance on each dataset than existing state of the art method. As we are using same backbone (OmniVec-4) for each of these datasets, it shows the robustness of the embeddings and the capacity of the network to adapt to different tasks and distribution of dataset.

\begin{table}[h]
\centering
\resizebox{\columnwidth}{!}{ 
\begin{tabular}{cccc}
\hline
\textbf{Dataset} & \textbf{Metric}  & \textbf{OmniVec} & \textbf{SOTA}        \\
\hline
AudioSet(A)      & mAP              &  54.8                & 53.3 (MAViL~\cite{huang2022mavil})         \\
AudioSet(A+V)    & mAP              &      55.2            & 51.2 (CAV-MAE~\cite{gong2022contrastive})          \\
SSv2             & Top-1 Acc   &    85.4              & 77.3 (MVD~\cite{wang2022masked})           \\
ImageNet1K       & Top-1 Acc   &      92.4            & 91.1 (BASIC-L~\cite{chen2023symbolic})       \\
Sun RGBD         & Top-1 Acc   &    74.6              & 67.2 (Omnivore~\cite{girdhar2022omnivore})      \\
ModelNet40       & Overall Acc &    96.6              & 95.4 (GeomGCNN~\cite{srivastava2021exploiting}) \\
\hline
\end{tabular}
}
\caption{\textbf{Comparison with state of the art} after fine tuning on respective training sets.}
\label{tab:sota}
\end{table}

\begin{figure*}[t]
    \centering
     \includegraphics[width=0.88\textwidth]{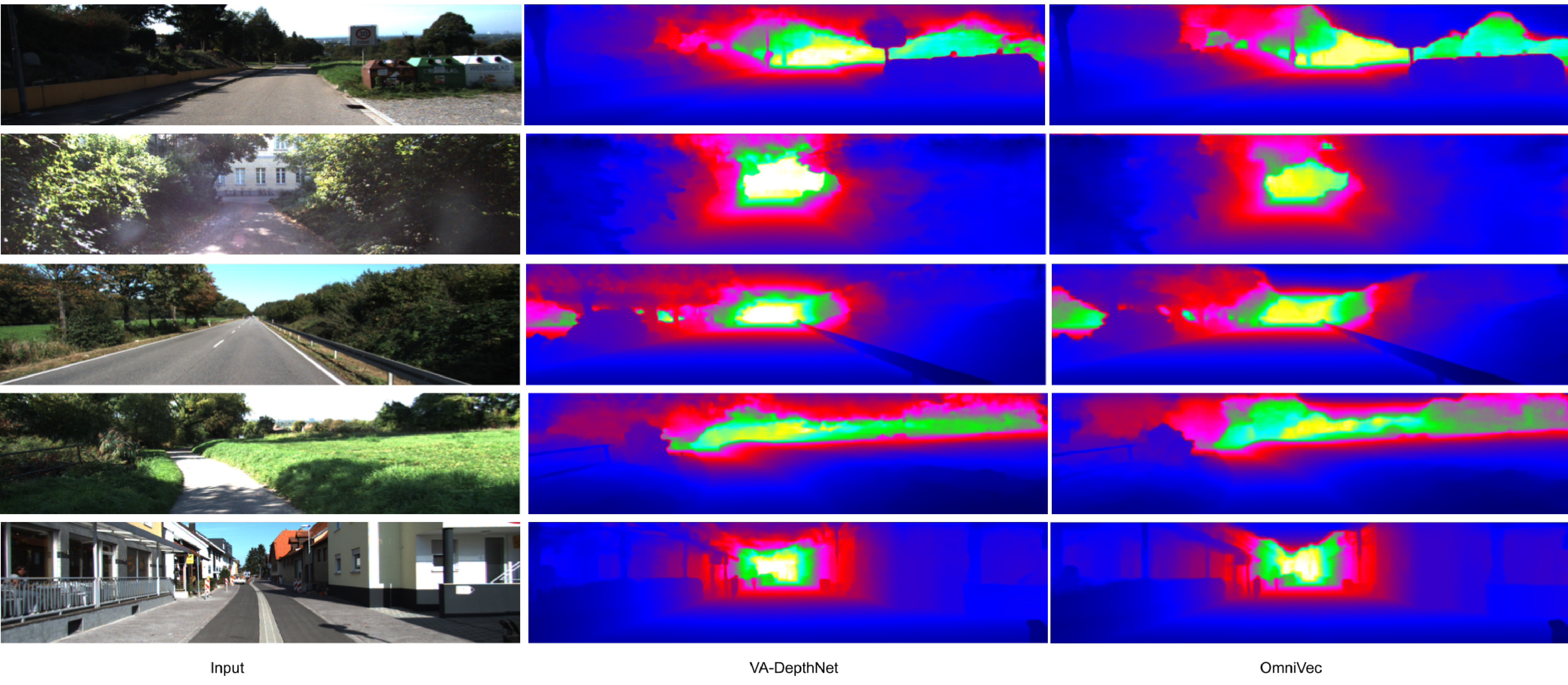}
    \caption{Qualitative results on test set of KITTI Depth Prediction. Ground truth is not available. For an RGB input image (left), the outputs from VA-DepthNet~\cite{liu2023va}(middle) and OmniVec (right) are shown. See supplementary material for more qual.~results.}
    \label{fig:kitti_qual}
\end{figure*}

\subsection{Generalization Ability}

\noindent\textbf{Generalization on unseen datasets.} We evaluate the performance of the learned embeddings on unseen datasets. Specifically, we show results on the tasks of fine grained image classification (Oxford-IIIT Pets\cite{parkhi2012cats}), Video Classification (UCF-101\cite{soomro2012ucf101}, HMDB51\cite{kuehne2011hmdb}), 3D point cloud classification (ScanObjectNN\cite{uy2019revisiting}), 3D point cloud segmentation (NYUv2\cite{silberman2012indoor}) and text summarization (SamSum\cite{gliwa2019samsum}). Our findings, tabulated in Table \ref{tab:generalization} [rows 1-6], demonstrates that even without fine-tuning, OmniVec surpasses most state-of-the-art methods. Further, while the pretrained OmniVec slightly underperformed on ScanObjectNN ($92.1$\%) compared to PointGPT's $93.4$\%, when fine-tuned, OmniVec achieved $96.1$\% accuracy, outperforming PointGPT. This shows OmniVec's generalizability on datasets where it is exposed to analogous tasks. 

\noindent\textbf{Generalization on unseen tasks - Monocular Depth Prediction on KITTI Depth Prediction Benchmark.} We fine tune the network for the task of depth prediction on KITTI Depth Prediction benchmark~\cite{Uhrig2017THREEDV}. Our network has not seen such image to image style transfer tasks. The results on KITTI depth prediction benchmark are shown in Table~\ref{tab:generalization} (row 7). We outperform the state of the method VA-DepthNet~\cite{liu2023va} \ie 10.44 iRMSE on VA-DepthNet \cf 10.2 for OmniVec. As can be observed from Figure~\ref{fig:kitti_qual}, the depth maps obtained by OmniVec are able to better capture the details near edges. 

\noindent\textbf{Cross-domain generalization.} Following prior work~\cite{akbari2021vatt}, we evaluate on the task of zero-shot text-to-video retrieval. 

The results are reported in Table \ref{tab:generalization}. On the YouCook2 dataset, our pretrained OmniVec surpasses the state of the art in zero-shot retrieval, achieving a Recall@10 of 64.2\% compared to VideoCLIP's 63.1\%. On MSR-VTT, when compared with SM~\cite{zeng2022socratic}, our fine-tuned OmniVec embeddings yield a Recall@10 of 89.4\% against SM's 90.8\%. With just pretraining, SM has a Recall@10 of 80\%, slightly above our 78.6\%. SM utilizes large-scale pretraining on internet scale data, while OmniVec uses much less data. Further, the second-best MSR-VTT method~\cite{chen2023vast} achieves only 73.9\% Recall@10 (see supplementary), which is behind our pretrained OmniVec.

%% file: conclusion.tex
\section{Conclusion and Limitations}
\label{sec:conclusion}
\noindent\textbf{Conclusion.} 

We proposed OmniVec, a unified data and task agnostic learning framework with a single backbone. The main idea behind OmniVec is that modalities in different domains can aid learning process. Further, we also proposed a novel training mechanism by grouping tasks and constructing mini batches by mixing inter-modality datasets. With experiments on $22$ datasets spanning across image, video, point cloud, depth, audio, text; we show that the proposed framework is highly generalizable along with being extremely robust. It can also generalize well to seen tasks with different data distribution as well as can adapt to unseen tasks effectively. We also studied the cross-domain knowledge sharing by evaluating a zero shot video-text retrieval task. We achieve state of the art or close to state of the art performance on all the evaluated datasets.

\noindent\textbf{Limitations.} 
OmniVec trains on unpaired multi-modal data, but paired data, though better, is expensive to obtain. The method employs multiple encoders per modality, increasing computational demands. Future research may address these computational challenges in unified networks.

\noindent\textbf{Societal Impact.} Modality agnostic techniques enhance realistic data cloning, risking misinformation and identity theft. These networks, syncing various modalities and using extensive internet data, amplify privacy, security, and bias concerns.

%% file: supp.tex
\section{Ablation on increasing number of parameters of base encoders} 
The details on influence of increasing the number of parameters termed as Modified encoder, of base modality encoders is provided in Table~\ref{tab:abl_size}. Our observations are as follows:

\begin{table*}[h]
\resizebox{\textwidth}{!}{ 

\begin{tabular}{ccccccc}
\hline
\textbf{Dataset} & \textbf{Metric} & \textbf{Modality Encoder} & \textbf{Base Encoder} & \textbf{Modified Encoder}   & \textbf{OmniVec (Pre.)} & \textbf{OmniVec (FT)}      \\
\hline
AudioSet(A)      & mAP              &  AST                & 48.5 & 49.4 & 44.7 & 54.8        \\
AudioSet(A+V)    & mAP              &      AST            & - & - & 48.6 & 55.2          \\
SSv2             & Top-1 Accuracy   &    ViViT             & 65.4 & 68.6 & 80.1 & 85.4      \\
ImageNet1K       & Top-1 Accuracy   &      ViT            & 88.5 & 89.1 & 88.6 & 92.4    \\
Sun RGBD         & Top-1 Accuracy   &    Simple3D-former              & 57.3 & 62.4 & 71.4 & 74.6    
\\
\hline
\end{tabular}
}
\caption{\textbf{Impact of increasing backbone size of base modality encoders.} All the base modality encoders above are based on ViT architecture. We increase the number of parameters equivalent to our OmniVec-4 model, by replicating the number of layers. }
\label{tab:abl_size}
\end{table*}

\noindent\textbf{OmniVec's Performance}: OmniVec (FT), which is OmniVec(Pre.) after fine-tuning, consistently outperforms the other methods across all datasets. This suggests that fine-tuning OmniVec is beneficial and leads to superior performance.
\\
\noindent\textbf{Base vs Modified Encoder}: The Modified Encoder generally performs better than the Base Encoder. While, the degree of improvement varies across datasets such as on datasets like Sun RGBD, we notice a substantial improvement of 5.1 percentage points, others like ImageNet1K and AudioSet(A) show relatively minor improvements. However, this relative improvement is significantly lower as compared to that obtained with OmniVec(Pre.) or OmniVec(FT). This suggests that the modifications may be especially beneficial for certain types of data or tasks, while training on multiple modalities provides consistent improvement across all modalities and tasks. This also indicates robustness and versatility achieved by OmniVec.

\begin{figure*}[ht]
    \centering
    \includegraphics[width=\textwidth]{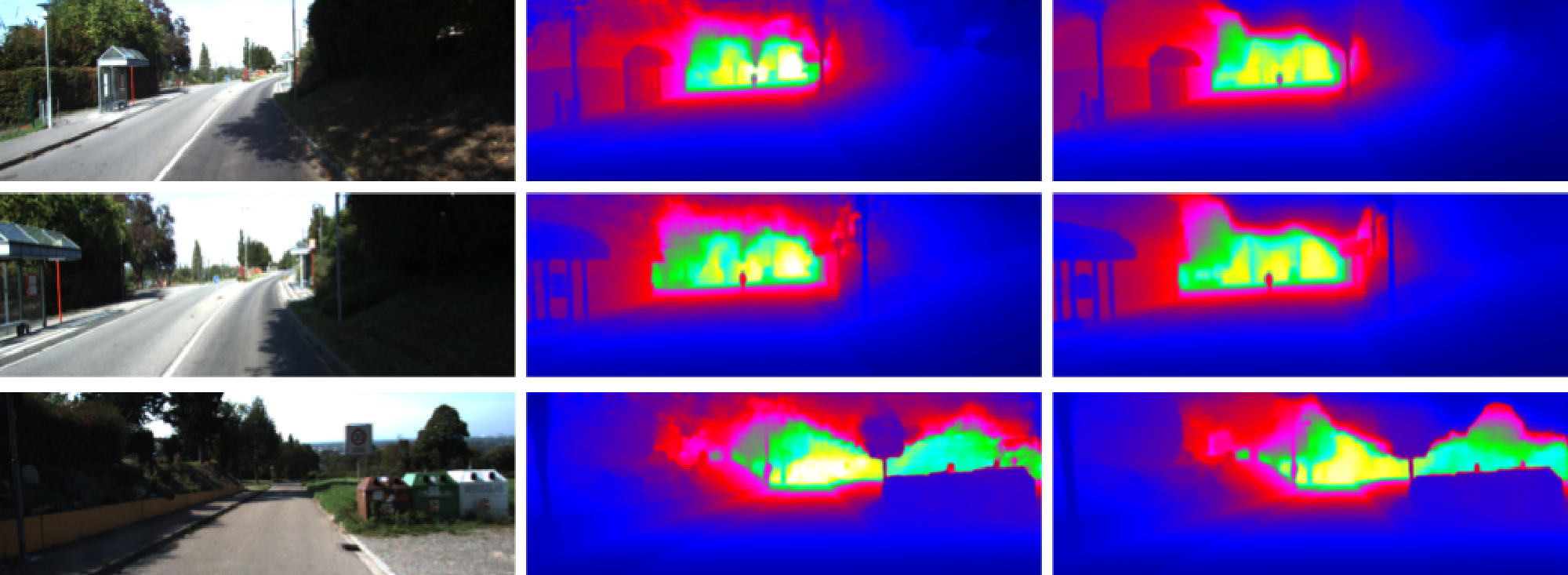}
    \caption{More qualitative results on Monocular depth prediction on KITTI test set. (From left to right) Input Image, Depth image generated using VA-DepthNet, Depth image generated using OmniVec. It can be observed that OmniVec predicts sharper depth around far away objects and on boundaries.}
    \label{fig:kittiqual}
\end{figure*}

\section{More Implementation Details} 
In addition to the datasets used for masked pretraining and training on multiple modalities, we also report results on additional datasets including both seen and unseen tasks. We use standard train/test split for each of the datasets for training and evaluating OmniVec \ie masked pretraining, training on multiple tasks, modalities and generalization. 

For demonstrating the generalization on unseen datasets, we compare the results against state-of-the-art methods on Oxford-IIIT Pets (image classification) \cite{parkhi2012cats}, UCF-101 \cite{soomro2012ucf101}, HMDB51 (video action recognition) \cite{kuehne2011hmdb}, ScanObjectNN (3D point cloud classification) \cite{uy2019revisiting}, NYU v2 seg (point cloud segmentation) \cite{silberman2012indoor} and SamSum (text summarization) \cite{gliwa2019samsum}. We evaluate the method on unseen task on KITTI depth prediction~\cite{Uhrig2017THREEDV}. We obtain results on standard test sets for each of the tasks. 

We do not fine-tune the base OmniVec network on any of these tasks and term it as OmniVec(Pre.) throughout the main manuscript (unless specified explicitly otherwise). The input to the network is the respective modality (text, image, point cloud, audio \etc). It is encoded with the respective encoders for these modalities as described in Table 1 (main manuscript) irrespective of the task. 
\\
\noindent\textbf{Segmentation and Summarization.} For segmentation and summarization, we use the same networks as described in Section 4-Task Heads (main manuscript). For reporting results with OmniVec(Pre.), we do not fine tune either encoder or decoder for evaluation on these tasks. 
\\
\textbf{Classification.} For classification/recognition tasks, as the classes differ from our training classes, following earlier works, we replace the Task Heads with a network consisting of two fully connected layers and a softmax classifier. We train these two layers by extracting OmniEmbeddings using the pretrained encoders of Table 1 (main manuscript) and the backbone Transformer network. We term it as OmniVec(Pre.) and we do not fine tune the backbone or the respective encoders to report results on it. For reporting results with fine-tuning (OmniVec(FT)), we use the pretrained OmniVec and fine-tune the network end-to-end on the respective training sets. 
\\
\noindent\textbf{Depth Prediction.} We use convolution decoder from \cite{ranftl2021vision} with our common transformer backbone. As the decoder works on patch wise output from the transformer encoder, we do not use a linear layer to reduce the features. We fine-tune the network in an end-to-end manner.
\\
\section{Detailed comparison with SoTA}

\noindent\textbf{Video Classification on UCF-101.} Table~\ref{tab:ucf} shows results on UCF-101 dataset for action recognition on $3$-fold accuracy. 

\begin{table*}[h]
\centering
\parbox{0.55\columnwidth}{
\centering
\begin{tabular}{ll}
\hline
\textbf{Method}     & \textbf{UCF-101}  \\
\hline
VATT~\cite{akbari2021vatt}       & 87.6  \\
Omnivore~\cite{girdhar2022omnivore}       & 98.2             \\
Text4Vis~\cite{wu2023revisiting}      & 98.2             \\
SMART~\cite{gowda2021smart} & { 98.6}              \\
VideoMAE V2-g~\cite{wang2023videomae} & \textbf{99.6}              \\
\hline
OmniVec(Pre.)    & {\ul 98.71}   \\     
OmniVec(FT)    & \textbf{99.6}   \\  
\hline
\end{tabular}
\caption{\textbf{UCF-101 Action Recognition}. Metric is 3-fold accuracy.}
\label{tab:ucf}
}
\hspace{1em}
\centering
\parbox{0.65\columnwidth}{
\centering
\begin{tabular}{m{4em}l}
\hline
\textbf{Method}     & \textbf{HMDB51}  \\
\hline
VATT~\cite{akbari2021vatt}       & 66.4  \\
DEEP-HAL~\cite{wang2021self}      & {87.56}            \\
VideoMAE V2-g~\cite{wang2023videomae}      & {88.10}            \\
\hline
OmniVec(\scriptsize{Pre})  & {\ul89.21}   \\   
OmniVec(\scriptsize{FT})    & \textbf{91.6}   \\   

\hline
\end{tabular}
\caption{\textbf{Comparison to state-of-the-art methods on HMDB51 dataset for Action Recognition}. Metric is 3-split accuracy.}
\label{tab:hmdb}
}
\hspace{1em}
\centering
\parbox{0.77\columnwidth}{
\centering
\begin{tabular}{lm{3em}m{3em}}
\hline
\textbf{Method}     & \textbf{Pets (top-1)} & \textbf{Pets (top-5)}  \\
\hline
Omnivore~\cite{girdhar2022omnivore}       & 95.1       &  {99.1}        \\
IELT~\cite{xu2023fine} & 95.28 & - \\
DINOv2\cite{oquab2023dinov2} & {96.70} &    -             \\
EffNet-L2~\cite{foret2020sharpness} & {97.10} &    -             \\
\hline
OmniVec(Pre.)    & {\ul 97.36}  & {\ul 99.3} \\          
OmniVec(FT)    & \textbf{99.2}  & \textbf{99.7} \\          

\hline

\end{tabular}
\caption{\textbf{Comparison to state-of-the-art methods on Fine grained image classification on Oxford-IIIT Pets dataset}. The metrics are top-1 and top-5 accuracy.}
\label{tab:pets}
}
\end{table*}

\noindent\textbf{Video Classification on HMDB51} Table \ref{tab:hmdb} shows comparison of state of the art method on HMDB51 dataset. 

\noindent\textbf{3D Point Cloud Classification on ScanObjectNN.} Table \ref{tab:scannet} compares OmniVec against state of the art methods. 

\begin{table*}[]
\centering
\parbox{0.55\columnwidth}{
\centering
\begin{tabular}{lm{4em}}
\hline
\textbf{Method}     & \textbf{Scan Object NN}  \\
\hline
PointConT~\cite{liu2023point}       & 90.3  \\
ReCon~\cite{qi2023contrast}      & {91.3}             \\
ULIP-2~\cite{xue2023ulip}      & {91.5}             \\
PointGPT\cite{chen2023pointgpt}      & {\ul 93.4}             \\
\hline
OmniVec(Pre.)    & {92.10}   \\        
OmniVec(FT.)    & \textbf{96.10}   \\          

\hline
\end{tabular}
\caption{\textbf{Comparison to state-of-the-art methods on ScanObjectNN for 3D point cloud classification}. Metric is Overall Accuracy.}
\label{tab:scannet}
} 
\hspace{1em}
\parbox{0.7\columnwidth}{
\centering
\begin{tabular}{ll}
\hline
\textbf{Method}     & \textbf{NYUv2}  \\
\hline
Omnivore~\cite{girdhar2022omnivore}       & 56.8  \\
CMN~\cite{liu2022cmx}      & {56.9}             \\
\hline
OmniVec(Pre.)    & {\ul 58.6}   \\   
OmniVec(FT)    & \textbf{60.8}   \\          

\hline
\end{tabular}
\caption{\textbf{Comparison to state-of-the-art methods on NYU v2 for semantic segmentation}. Metric is mean IoU. Note that the network has not been fine-tuned on this dataset nor any additional network has been used.}
\label{tab:nyu}
} 
\hspace{1em}
\parbox{0.72\columnwidth}{
\centering
\begin{tabular}{ccccc}
\hline
\textbf{Method} & \textbf{R-1}   & \textbf{R-2}   & \textbf{R-L}    \\
\hline
Pegasus~\cite{zhao2022calibrating}       & 54.37          & 29.88          & 45.89                 \\
MoCa~\cite{zhang2022momentum}            & {\ul 55.13}          & \textbf{30.57}          & {50.88}                 \\
\hline
OmniVec(Pre.)         & {54.81} & {\ul 30.10} & {\ul 51.21}  \\
OmniVec(FT)         & \textbf{58.81} & \textbf{31.1} & \textbf{53.4}  \\

\hline
\end{tabular}
\caption{\textbf{SamSum dataset for meeting summarization}. Metric are ROGUE scores. Note that the network has not been fine-tuned on this dataset nor any additional network has been used.}
\label{tab:samsum}
}
\end{table*}

\noindent\textbf{Semantic Segmentation on NYU v2 seg.} Table \ref{tab:nyu} shows result on NYU v2 segmentation against state of the art methods.

\noindent\textbf{Text summarization on SamSum.} Table~\ref{tab:samsum} compares our method against state of the art methods. An important observation here is that our method has not been specifically trained for solving text related tasks nor the OmniVec(Pre.) network has been fine tuned for this dataset. This shows that the learning mechanism is able to generalize across domains as well.

\section{Qualitative Results.} We present additional qualitative findings in Figure~\ref{fig:kittiqual}. When compared to the previous benchmark, VA-DepthNet, our suggested approach demonstrates superior depth perception at boundaries and distant objects. Notably, our method offers enhanced depth discernment for objects such as the bus shelter (highlighted in the top and middle images) as well as houses (depicted in the bottom image).